\documentclass[letterpaper, 10 pt, conference]{ieeeconf}  % Comment this line out if you need a4paper
\usepackage{graphicx}

\IEEEoverridecommandlockouts                              % This command is only needed if 
\title{\LARGE \bf
Learning View and Target Invariant Visual Servoing for Navigation
}
\author{Yimeng Li and Jana Ko\v{s}ecka \thanks{George Mason University, Computer Science Department {\tt\small [yli44, kosecka]@gmu.edu}}}

\begin{document}

\maketitle
\thispagestyle{empty}
\pagestyle{empty}

%%%%%%%%%%%%%%%%%%%%%%%%%%%%%%%%%%%%%%%%%%%%%%%%%%%%%%%%%%%%%%%%%%%%%%%%%%%%%%%%
\begin{abstract}
The advances in deep reinforcement learning recently revived interest in data-driven learning based approaches to navigation.
% An important navigation skill for view based navigation is the ability to reach the desired view or navigate towards the target of interest. Both of these tasks have been tackled in the past using traditional geometric visual servoing methods.  
In this paper we propose to learn viewpoint invariant and target invariant visual servoing for local mobile robot navigation; given an initial view and the goal view or an image of a target, we train deep convolutional network controller to reach the desired goal. %  Additional advantage of the proposed approach is that it does not require accurate localization.
We present a new architecture for this task which rests on the ability of establishing correspondences between the initial and goal view and novel reward structure motivated by the traditional feedback control error. The advantage of the proposed model is that it does not require calibration and depth information and achieves robust visual servoing in a variety of environments and targets without any parameter fine tuning. We present comprehensive evaluation of the approach and comparison with other deep learning architectures as well as classical visual servoing methods in visually realistic simulation environment~\cite{xia2018gibson}.
The presented model overcomes the brittleness of classical visual servoing based methods and achieves significantly higher generalization capability compared to the previous learning approaches.

% demonstrating that the ability of establishing image correspondences is central to many tasks. 

% {\em mention holonomic /non-holonomic, pose metrics}

\noindent
% Start from Vladlen Koltun's idea. If we train the representation from scratch, like locomotionNet or learn to see by moving. I tried a lot but these representations doesn't work very well.

% But classical methods seem to be working no matter how large the distance is. 

% So if we take the correspondence idea from classical method, classical method also has inconsistent action when there are too many correspondences.

% So we take the idea from learning method to deal with the dense correspondence problem.

% Comparing learning method with classical method through many correspondences. Classical method cannot deal with it. And also without depth.

% Comparing different learning methods through using different visual representaiton.

% Also mention different pose metrics I'm using. Without angle, without alpha, with alpha and beta.

\end{abstract}

%%%%%%%%%%%%%%%%%%%%%%%%%%%%%%%%%%%%%%%%%%%%%%%%%%%%%%%%%%%%%%%%%%%%%%%%%%%%%%%%
\section{INTRODUCTION}

% While the traditional navigation architectures required integration of mapping, motion planning and low level control, several approaches emerged which learn various navigation skills as direct mappings from raw observations to actions. 

% Biologically motivated approaches outline two fundamentally different spatial behavior groups: local navigation and way-finding. 

Traditional approaches to navigation in novel environments often required solution to many components, including
mapping, motion planning and low level control. The reliance of motion planning on high quality geometric maps and trajectory following on perfect localization, resulted in fragmented and brittle solutions which had to be fine-tuned for particular environments. % Mapping was focus of large body of work on simultaneous localization and mapping SLAM~\cite{} focusing on the construction of metric or topological maps. 
In contrast to these methods biological systems 
have more flexible representations of environments and control policies which enable them to robustly navigate in previously unseen environments. These observations and the emergence of effective data driven techniques for learning control policies directly from observations recently spurred large body of research in learning  navigation. 
% Visual Servoing (VS) techniques have been applied to a variety of robotic tasks, including reaching, docking and navigation.
In this paper we propose a learning approach to visual servoing for mobile robots in indoors environments. In the broader context of navigation task, visual servoing discussed here can be viewed as local navigation skill for view based navigation, where the goal is to reach the desired view or navigate towards the target of interest. While there is a large body of work on classical approaches to visual servoing, they rely on the extraction, tracking and matching of a set of visual features which are difficult and brittle tasks. Related attempts to overcome these difficulties using learning based approaches have been recently considered in~\cite{bateux2017visual}, with the focus on improving the pose estimation and perception component for 6-DOF pose based visual servoing in table top environments. We study the problem in mobile robot navigation setting and propose to learn the control policy jointly with the perception component. 
The contributions of this work are as follows: \\
\noindent
%(i) We present a new architecture for image based visual servoing which can be trained end-to-end and demonstrate its generalization ability to both target reaching and goal view reaching task;
(i) We show how general image correspondence map can be distilled by deep convolutional networks and trained in an end-to-end manner to learn a policy for target reaching and goal view reaching task;
%(ii) We demonstrate how to train the model using deep reinforcement learning (DRL) framework with novel dense reward structure; 
(ii) We design a novel dense reward structure and train the model using deep reinforcement learning (DRL) framework;
(iii) We present a comprehensive comparison of our model with alternative deep CNN architectures and training approaches proposed for this task as well as classical image based visual servoing techniques, showing superior performance of our approach. \\
Comprehensive evaluation is carried out using visually realistic  simulated household environments~\cite{xia2018gibson} with variety of targets and goals, demonstrating good generalization ability of the approach.

\begin{figure}[t]
\centering
\includegraphics[width=0.48\textwidth]{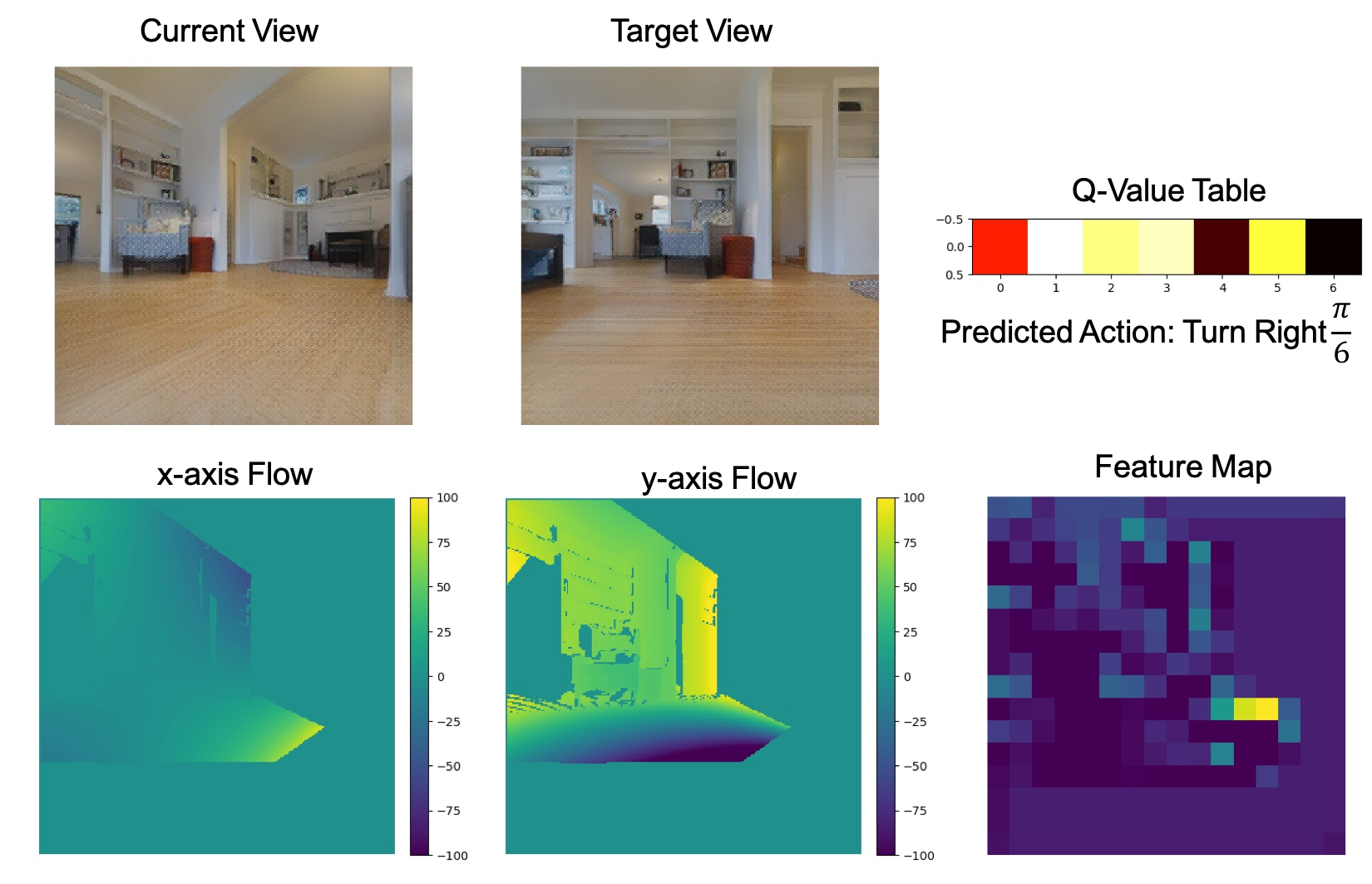}
\caption{Visualization of the proposed method. The three images on the top are current view, target view and Q-value table. Brightest color shows up in the left block of the Q-value table indicates taking a turning-right action. The three images at bottom are the visualizations of dense correspondence map on $x/y$-axis and the feature map output by the perception module.}
\label{fig:method introduction}
\end{figure}

\section{RELATED WORK}

%There is a large body of work on using learning based approaches to navigation. 
Here we review the related work focusing on local navigation skills most relevant to our approach. 
% Early successes of (deep) learning based approaches exploited the ability to learn features of the environment for mapping and localization considered traditional metric grid based representations~\cite{henriques2018mapnet, Chaplot2019neuralmap, gordon2018iqa}.
% Several works consider navigation tasks without explicit map representation. Many of these methods are loosely motivated by various naviagtion skills observed in biological systems~\cite{franz2000biomimetic}
A class of local navigation methods assumes that the goal is specified in an agent's local coordinate frame, often assuming perfect localization. Authors in~\cite{gupta2017cognitive} learn a local navigation policy using deep reinforcement learning and Value iteration networks,~\cite{bansal2019combining} learn how to predict the next waypoint given a long-range goal and use traditional optimal control to compute the desired trajectory given local ego-centric map of the environment.  
Kumar et al.~\cite{kumar2018visual} consider visual-teach-and-repeat approach  where the environment is represented in terms of trajectories experienced in exploration. Efficient retrieval of the views along with the actions enables novel traversals of the environment. 

% learning based visual-teach-and-repeat
% The class of locomotion policies where initial and goal view are specified, are typically trained to predict the desired motion using egomotion and vary in the proposed architecture, loss function and evaluation methodology. 

Locomotion policies proposed in~\cite{agrawal2015learning} and~\cite{jayaraman2017learning} use siamese network to extract feature maps from the input images and estimate the discrete motion. Pathak et al.~\cite{pathak2018zero} combines forward dynamics with inverse dynamics model to solve the bi-modal ego-motion estimation problem. Savinov et al.~\cite{savinov2018semi} stack the start and goal image up as an input to their locomotion network, training the model in a supervised way. Disadvantage of these models is poor generalization capability and very dense sampling of the intermediate views to represent the trajectories. 

%Many recent works use deep re forcement learning (DRL) to achieve long-range navigatiinn. Target representations vary a lot. Bansal et al.~\cite{bansal2019combining} and Mirowski et al.~\cite{mirowski2018learning} tackles the goToPoint task. Bensal uses a perception network to predict the next waypoint given the long-range goal represented as location coordinates while Mirowski directly represents the goal as the distance to other places. Kumar et al.~\cite{kumar2018visual} saves all the views along the visited trajectory efficiently in memory for the 'homing'~\cite{franz2000biomimetic} behavior. Bruce et al.~\cite{bruce2018learning} uses DRL to navigate through pre-saved panoramic images and reach an image-represented goal. Faust et al.~\cite{faust2018prm} build a probabilistic roadmap to achieve long-range navigation and use DDPG for continuous action space. 
Related problem of semantic target driven navigation was  considered by~\cite{sadeghi2019divis,zhu2017target} where object is specified as an image or as an semantic category~\cite{mousavian2019visual} % In~\cite{mousavian2019visual} and~\cite{zhu2017target} authors did not assume the object was initially in the FOV of the agent, 
considering mid-range navigation tasks  and more loose definition of goal success. Similarly to us many of the learning based methods train their models in simulated environments, followed by transfer or adaptation of the learned policies to real robots. 
% There are also plenty of works utilize simulated environment to achieve virtual-to-real transferring. 
% Mousavian et al.~\cite{mousavian2019visual} and Wortzman et al.~\cite{wortsman2019learning} both train their navigator in a simulated environment and tackle the GoToObject task. 
Authors in~\cite{meng2019neural} and~\cite{gupta2017cognitive} successfully transfer their model trained on GibsonEnv~\cite{xia2018gibson} to the real world. 
% Gordon et al~\cite{gordon2019splitnet} and Hong et al.~\cite{hong2018virtual} exploit intermediate representations including surface normal, depth or semantic segmentation representations to generalize their model trained on simulated data to the real world. 
% add for review2

Besides going to a semantic goal, people have also explored other goal representations;  Amini et al.~\cite{amini2019variational} use a local topological map 
% as a direction guide and successfully deploy the trained policy on a real Toyota vehicle. 
and Codevilla et al.~\cite{codevilla2018end} steer a toy truck through high-level map related commands.

% In terms of short-range navigation, people tend to predict the ego-motion between a pair of images through supervised learning. In \cite{agrawal2015learning} and ~\cite{jayaraman2017learning}, they use to siamese network to extract feature maps from the input images and estimate the discrete motion. Pathak et al.~\cite{pathak2018zero} combines a forward dynamics with inverse dynamics model to solve the bi-modal ego-motion estimation problem. Savinov et al.~\cite{savinov2018semi} stacks two images up and predict actions under supervision.

There is also a large body of work on classical visual servoing methods. More recent discussion can be found in~\cite{corke2017robotics}.  Image based visual servoing  has been adapted for short range  mobile robots navigation in~\cite{bista2016indoor} and pose based visual servoing task was studied in~\cite{bateux2017visual} where 6-DOF pose regression is 
estimated by deep networks followed by traditional control.

The work closest to ours is Sadeghi et al.~\cite{sadeghi2019divis, sadeghi2018sim2real} and Yexin et al.~\cite{ye2019gaple}. They both study the target reaching problem and use DRL to train the model in simulation. In~\cite{sadeghi2019divis} the policy is guided by heatmap obtained through the correlation in the feature spaces of object  and current robot's view. The attention mask obtained from semantic segmentation is computed by~\cite{ye2019gaple} and used as an input for the policy learning. Both approaches focus on the problem of reaching {\em semantic target} and rest on the availability of powerful architectures and representations pre-trained for object detection and semantic segmentation.
Furthermore the attention mechanism does not lend itself to {\em homing} scenarios in navigation, where the goal of the agent is specified as the goal view and may not contain interesting objects.

\section{APPROACH}

Consider an agent that operates in a novel indoor environment. Instead of navigating with the help of a pre-built metric map, our agent is provided with a set of images taken at different locations, forming a view based topological map of the environment~\cite{savinov2018semi}. Our goal is to train the agent to do short-range navigation to reach desired location or target of interest in the field of view of the agent. Let's assume that the agent starts at some random state $s_0$ and obtains an observation $I_0$. State $s$ is characterized by the pose of the agent $x, y$ coordinates and the heading angle $\theta$ in a world reference frame unknown to the agent. 
The observation $I$ can be either an RGB image or an RGB-D image including depth information as additional channel. A target view image $I_g$ taken at goal position $s_g$ is provided to the agent. $I_g$ is assumed to have some overlap with the initial field of view $I_0$. The goal for the agent is to learn a policy for reaching the goal state $s_g$ by executing a sequence of actions $a$. The action space $A$ can be either continuous or discrete depending on the control method.

% We tackle the GoToView problem in two approaches:
We approach this problem using learned data-driven strategy and compare and discuss the advantages of this approach to classical image based visual servoing~\cite{hutchinson1996tutorial} and learned visual servoing approaches~\cite{bateux2017visual}. 

Classical visual servoing computes correspondences at few selected points between current observation $I_t$ and target view $I_g$ followed by analytic derivation of the feedback control law for continuous velocities at each each step. Learning method learns a mapping from  $I_t$ and $I_g$ to actions in reinforcement learning framework and predicts at each state a discrete action $a$ moving towards the goal location. The action space includes seven actions, which are moving forward $0.1m$ in 7 orientations $[-\frac{\pi}{4}, -\frac{\pi}{6}, -\frac{\pi}{15}, 0, \frac{\pi}{15}, \frac{\pi}{6}, \frac{\pi}{4}]$.\\

\noindent
We will start our discussion with the geometry of the problem, discussing the brittleness and drawbacks of the geometric methods
to motivate our learning based solution. 

\subsection{Classical Visual Servoing}
 Visual servoing control has been developed in early 90's with the goal of increasing the accuracy of the control of robot end effector by using visual feedback control. Two main classes of systems are position based and image based visual servo. While the position based visual servo strives to first estimate the relative pose between the initial and target view, the image based visual servo derives the error signal directly from measurements.
 The image features $f$ can vary dramatically (e.g. points or lines), but the effective relationships between robot's pose and its velocities is characterized by feature Jacobian $J_f(s)$ which can be derived analytically
 \begin{equation}
 %\[ \dot f = J_f({\bf s}) \dot{{\bf s}} \] 
 \label{fundamental_vs_equation}
 \dot f = J_f({\bf s}) \dot{{\bf s}}
 \end{equation}
  where $\dot{{\bf s}} = [{\nu_{x}}, {\nu_{y}}, {\nu_{z}} , {\omega_{x}} , {\omega_{y}} , {\omega_{z}}]^T$. $J_f(s)$ is also referred to as {\em interaction matrix}. 
% Image-Based Visual Servo (IBVS) is a control technology that generate actions to move feature points $p$ in current image to the target points $p^*$ indicated by the goal image~\cite{corke2017robotics}. 
% Given points $p$ in current image and their corresponding coordinate in the goal frame $p^*$ the image error was $(p^*-p)$ denoted was $\dot{u}$ and $\dot{v}$. 
For image points the relationship between image velocities and motion of the end effector is characterized by the well known optical flow equation~\cite{corke2017robotics}.
% We do not need to estimate the relative pose between the two images as moving feature points in the robot's observation implicitly change the pose. We use an interaction matrix $L$ to associate camera velocity with feature velocity in image coordinates.
% \begin{equation}
% \label{interaction_matrix}
%\left[\begin{array}{c}{\dot{u}} \\ {\dot{v}}\end{array}\right]=J\left[\begin{array}{ccccccc}{\nu_{x}} & {\nu_{y}} & {\nu_{z}} & {\omega_{x}} & {\omega_{y}} & {\omega_{z}}\end{array}\right]^{T}
%\end{equation}
For a holonomic mobile robot moving on the xz-plane, the velocity space is three-dimensional $[\nu_x, \nu_z, \omega_{y}]^T$ and the optical flow equation reduces to:  
\begin{equation}
{\small 
\label{full_interaction_matrix}
\left[\begin{array}{c}{\dot{u}} \\ {\dot{v}}\end{array}\right]=\left[\begin{array}{ccc}{-\frac{\lambda}{Z}} & {\frac{\left(u-u_{0}\right)}{Z}} & {-\left(\lambda+\frac{\left(u-u_{0}\right)^{2}}{\lambda}\right)} \\ {0} & {\frac{\left(v-v_{0}\right)}{Z}} & {\frac{-\left(u-u_{0}\right)\left(v-v_{0}\right)}{\lambda}}\end{array}\right]\left[\begin{array}{c}{\nu_{x}} \\ {\nu_{z}} \\ {\omega_{y}}\end{array}\right]
}
\end{equation}
where $\dot{u}$ and $\dot{v}$ is the optical flow,
$\lambda$ is the focal length for the camera, $Z$ is the depth of the feature point and $(u_0, v_0)$ is the image principal point. 
In practice, if $N >= 2$ correspondences are detected, we can stack up rows $J_f$ to get the interaction matrix for all features $J$. The control law can then be obtained by computing the pseudo-inverse of Eq.~(\ref{full_interaction_matrix}) for the desired camera motion $\dot s^* = J_f^{+} (f^*-f)$. 
%\begin{equation}
%\left[\begin{array}{c}{\nu_{x}} \\ {\nu_{z}} \\ %{\omega_{y}}\end{array}\right]=\left[\begin{array}{c}{L_{1}} \\ %{\vdots} \\ {L_{N}}\end{array}\right]^{+}\left[\begin{array}{c}{\%dot{u}_{1}} \\ {\dot{v}_{1}} \\ {\vdots} \\ {\dot{u}_{N}} \\ %{\dot{v}_{N}}\end{array}\right]
%\end{equation}
% In terms of a non-holonomic mobile robot, it cannot move along the x-axis of its body frame. So we further reduce its velocity space into a two-dimensional vector $(\nu, \omega)$ through a transformation~(\ref{velocity_transformation}) from the camera velocity to the robot velocity.
% \begin{equation}
% \label{velocity_transformation}
% \begin{array}{l}
% \nu=\sqrt{v_{x}^{2}+v_{y}^{2}}\\
% \omega=\tan ^{-1} \frac{v_{y}}{v_{x}}+\omega_{y}
% \end{array}
% \end{equation}

For image based visual servoing (IBVS), in addition to challenges of detection, matching and tracking of geometric features, there are additional well-known difficulties for visual navigation tasks~\cite{corke2017robotics}. 
%\begin{itemize}
    %\item 
    It is possible to have an inconsistent set of feature velocities such that no possible camera motion will result in the required image motion. For example, nearly collinear features will cause very small camera motion.
    %\item 
    The performance of the model relies on the feature depth $Z$ in Equation~(\ref{full_interaction_matrix}). Some approaches considered using a constant depth for all the feature points or depth estimates from motion is also helpful. The motion based methods often failed when camera motion 
    got smaller.
    %\item Dense correspondence is not applicable to IBVS as it's computational expensive to find the pseudo-inverse for a large interaction matrix. And it is more likely that the feature velocities are inconsistent when you have a large number of features.
    %\item 
    For the configurations where the desired orientation is notably different from the current orientation and differential drive robots with non-holonomic constraints, the overlap between current view and goal view often becomes too small resulting in the failure of the correspondence computation. 
%\end{itemize}
The proposed learning based approach helps to overcome the brittleness of the traditional methods. 

\subsection{Learned Visual Servoing}
We formulate the problem of learning visual servoing (LVS) in the reinforcement learning framework. An agent starts at a state $s_0$ in the environment and the goal is to navigate to the given target view. At each $s_t$,  the agent receives an observation $I_t$. Given the observation, the agent takes action $a_t$ and receives a reward $r(s_t, a_t)$. 
We are interested in learning control policy 
$a_t = \phi_{\theta}(\psi(I_t, I_g))$ and representation of current view and goal view, that predicts the action advancing the agent towards the goal, where $\phi_{\theta}$ is the action module and $\psi$ is the perception module. In traditional Q-learning approach, the goal is to learn a Q-function $Q(s_t, a_t)$ quantifying the goodness of different actions in each state.  By following a policy $\pi_{\theta}$, the agent produces a sequence of Q-states $(s_t, a_t)^{T}_{t=0}$ after $T$ steps. Our goal is to train a policy $\pi_{\theta}$ that maximizes the total reward received through the trajectory. 

\begin{figure}[t!]
\centering
\includegraphics[width=0.48\textwidth]{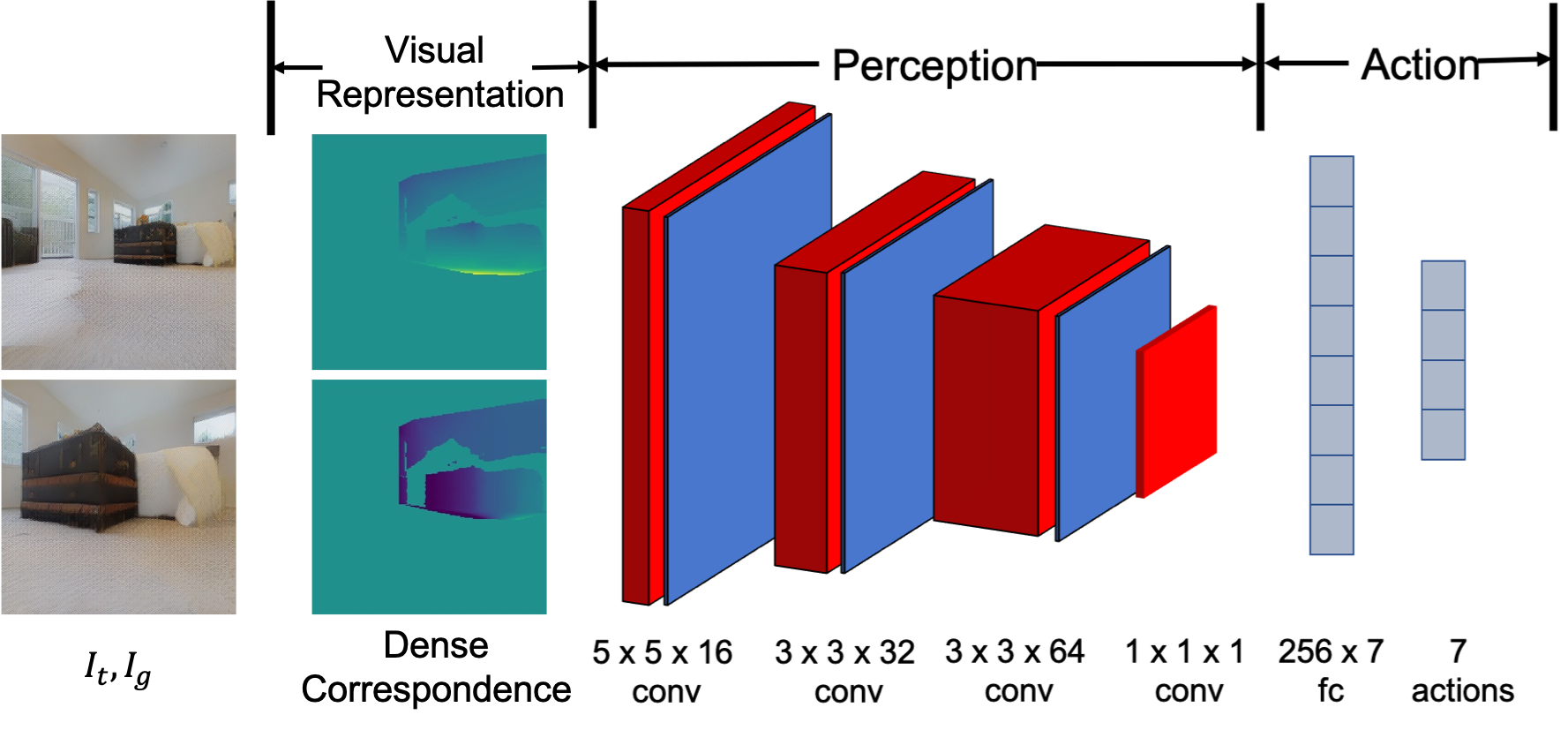}
\caption{Learned Visual Servoing Architecture. Visual representation is the dense correspondence extracted from two input images $I_t, I_g$. Perception module extracts features from the visual representation. Action module predicts the Q-values for each action based on the feature input.}
\label{fig:dqn_model}
\end{figure}

\noindent \textbf{Deep Q Network:}
For large state spaces the Q function cannot be computed exactly. 
We use a Deep Q Network (DQN) ~\cite{mnih2013playing} to approximate the value of a Q-state, $Q(s_t, a_t)$, the action-value of unobserved pose $s_t$ of the mobile agent.
Our DQN architecture has two components: a perception module and an action module as shown in Figure~\ref{fig:dqn_model}. The perception module consists of four convolutional layers, all using ReLU activation and batch normalization for efficient training. The first three convolutional layers have kernel size $5 \times 5 \times 16, 3 \times 3 \times 32, 3 \times 3 \times 64$, strides of $2, 1, 1$ and each of them is followed by a max pooling layer to reduce spatial support of the feature map. The last CNN layer is a $1 \times 1$ convolutional layer which outputs a $16 \times 16 \times 1$-sized feature map. As input we examine several intermediate representations of RGB views before passing them into the perception module. For the action component the feature map is flattened into a 256-dimensional vector followed by a fully-connected layer which outputs Q-values for all actions in our feedback policy (see Figure~\ref{fig:method introduction}). Note that our model's architecture is highly flexible. Given different kinds of visual input, we can vary the design of the perception module and leave the action module untouched. %Or we can with ease insert an LSTM layer in front of the fc-layer to perform as a recurrent policy.
% for review 2
We also tried to insert an LSTM layer in front of the FC-layer to learn a recurrent policy. We didn't get a performance boost compared to the reactive policy. All the experiment results mentioned in the paper are achieved without using an LSTM layer.

\noindent \textbf{Input Visual Representations:}
Motivated by classical visual servoing, the input to our perception module is the map of dense correspondences.
% However, because of the computational bottleneck of the interaction matrix used by classical visual servoing, we have to use sparse feature points after carefully feature selection instead of using all of them. In the knowledge of CNN's capability of dealing with large amount of data, we used correspondence map as the input to the DQN. 
In the first stage we compute the dense correspondences (shown in Figure~\ref{fig:dqn_model})
using ground truth information of known pose and camera depth.  
Learned and fine-tuned dense correspondences as in~\cite{choy2016universal} can be later incorporated in our model. The input is a two-channel image where each channel represents the relative offset on $x/y$-axis from each point in the current view to a corresponding point in the target view. We also observe that applying image smoothing to the dense correspondence map during testing stage improves the model's stability, as smoothed correspondence map has fewer depth  discontinuities that can negatively affecting the final prediction.
In experiments (Exp D) we corrupt the high-quality dense correspondences in various ways to see the effect of the noise on the final robustness of our model.

\noindent \textbf{Reward:}
For the task of driving to a target view, the robot traverses a trajectory so as to minimize the error between start view and goal view. Since view difference between two different locations with similar orientation is rather small, the image error is a weak signal for our learning task. To measure the 'progress' to the goal during navigation, we compute $d_{polar}$ distance between poses $(x_t, y_t, \theta_t)$ and $(x_g, y_g, \theta_g)$. $d_{polar}$ is expressed in the coordinate frame of the goal, where $\rho$ is the distance between robot's center and the goal position, $\alpha$ is the angle between robot's reference frame and the vector connecting center of the robot with the goal position and $\beta$ is the angle difference between current orientation of the robot and heading direction. The distance to the goal $d$ is a weighted sum of these position and angular distances. Note that when at the goal all $\rho = \alpha = \beta = 0$. The distance to the goal is related to feedback control law error for differential drive robots outlined in~\cite{corke2017robotics}.
\begin{equation}
\label{PolarPose metric}
\begin{array}{l}
\rho  =  \sqrt{(x_g - x_t)^2+(y_g - y_t)^2}\\
\alpha = \arctan(y_g-y_t, x_g-x_t) - \theta_t\\
\beta  = \theta_g - \arctan(y_g-y_t, x_g-x_t)\\
d_{polar}  = \rho+\lambda_{\alpha}\rho(\alpha, 0) + \lambda_{\beta}\rho(\beta, 0)
\end{array}
\end{equation}
$d_{polar}$ distance enforces the agent to travel towards the target  and rotate to align with target pose when the agent is in the vicinity of the target. In practice, $\lambda_{\alpha}$ and $\lambda_{\beta}$ are both set to 0.2. We also consider $d_{pose}$ distance~\cite{lavalle2006planning} for comparison and use $d$ for clarity to denote the distance below.

The reward is designed to encourage the robot to move in the direction to minimize the distance to the goal. The reward function is,
\begin{equation}
\label{equation_reward_setup}
R=\max \left(0,\min \left(\frac{d_{t-1}}{d_{init}}, 1\right)-\min \left(\frac{d_{t}}{d_{init}}, 1\right)\right)
\end{equation}
where $d_{init}$ is the initial distance between the agent and the goal and $d_{t}$ is the current distance of the agent to the goal. The right term in the $\max$ function measures the progress of the agent towards the goal after taking action $a_t$.

% for reviewer 2
Note that we are having a dense reward setup as the environment is simulated. We leave the design of sparse rewards to future work when we train an agent in real-world scenes.

\section{EXPERIMENTS}

\begin{figure}[t!]
\centering
\includegraphics[width=0.46\textwidth]{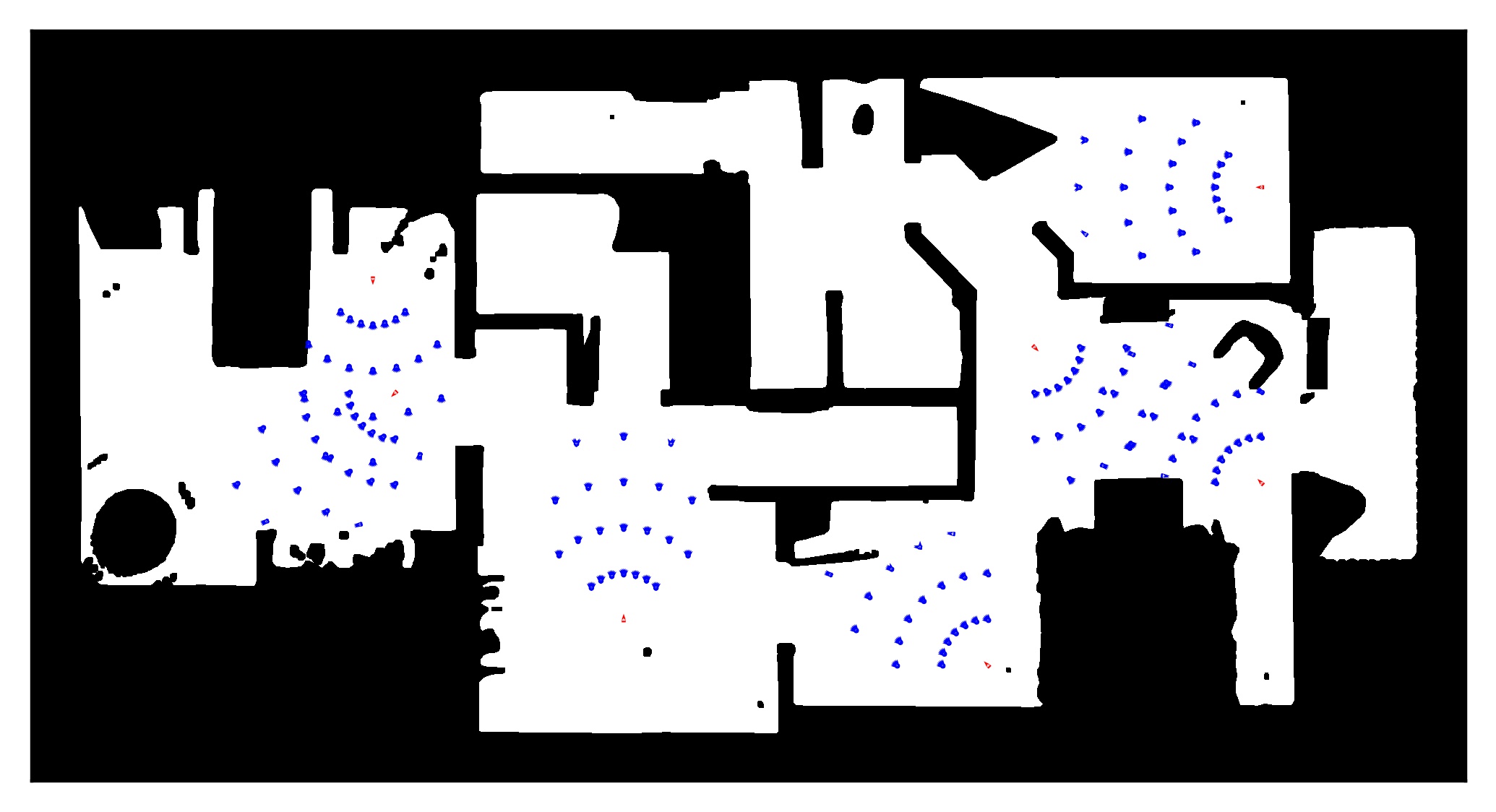}
\caption{Occupancy map of the 'Dardan' scene in the testing dataset. We sample 7 starting viewpoints in the environment represented by the red triangles. The blue triangles are the target poses. Their distance to the start viewpoint and orientation vary from each other.}
\label{fig:environment_map_and_poses}
\end{figure}

\begin{figure}[t!]
\centering
\includegraphics[width=0.48\textwidth]{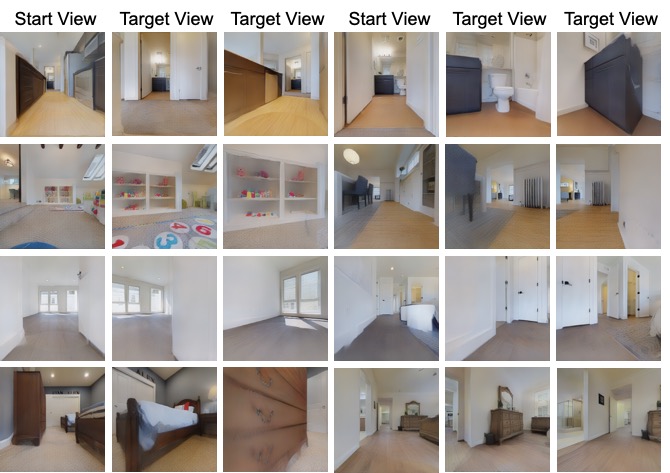}
\caption{Visualizations of start views and target images from different test scenes. First and fourth column shows the start views while second, third, fifth and sixth column shows the target views.}
\label{fig:environment_images}
\end{figure}

\noindent \textbf{Datasets}:
All the experiments are completed in GibsonEnv~\cite{xia2018gibson}. Gibson environment contains visually realistic reconstructions of indoors scenes, with varying appearances and layouts.  We randomly select 16 indoor scenes as training data and 6 others for testing. During training, we randomly sample the starting robot poses. Examples of sampling on one test scene is shown by Figure~\ref{fig:environment_map_and_poses}. Sampled initial views and target views are shown by Figure~\ref{fig:environment_images}. The target views are chosen at certain distance and orientations away from the starting point. The distance ranges from $0.5$ to $4.0m$. The orientation and target pose heading angle ranges from $-\pi/4$ to $\pi/4$ with respect to the initial pose. 
% add for reviewer 2
Usually it will take the agent from 5 to 35 steps to reach the goal.
We make sure that the target view and start view has at least $16 \times 16$ pixel overlapping area. During testing, for each scene, we manually select approximately 10 starting poses where there is enough open space in front of them. The target images are chosen in a similar way as training data. We evaluate different approaches by computing the success rate of visual servoing results. The success criteria is that the robot pose is within $0.2m$ to the target pose during navigation. 

\noindent \textbf{Training Details}:
We use Actor-Critic architecture~\cite{mnih2013playing} and replay memory when training our model. Each minibatch consists of losses over 128 random state-action tuples sampled from replay memory. Each tuple contains the state representation, the action and the reward after taking the action. We use RMSprop Optimizer, learning rate of $10^{-5}$. The model is optimized for $50000$ iterations on a single GPU, which takes 6 hours.

\subsection{Intermediate Representations for LVS}
We compare the performance of different perception module architectures while using the same reward structure and action module in DQN.

\begin{figure*}[t!]
\centering
\includegraphics[width=0.94\textwidth]{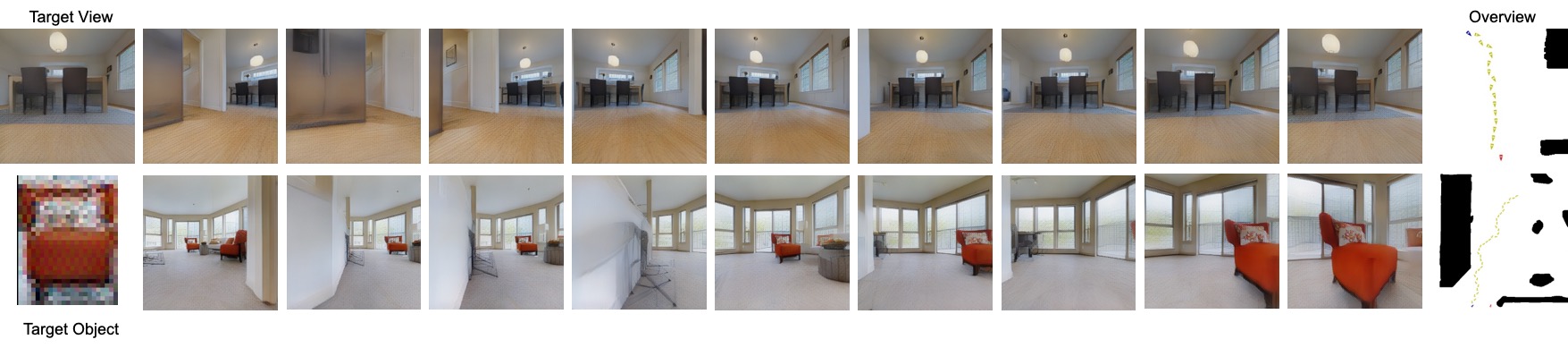}
\caption{Examples of a robot driving to a view (first row) and driving to an object (second row). The first column shows the target view and the target object. The last column is the overview of the two trials. Between them are the observations along the driving.}
\label{fig:GoToView_and_GoToObj}
\end{figure*}

\noindent\textbf{LocomotionNet}~\cite{savinov2018semi}. Input to the network is a triplet of RGB images comprising of previous observation, current observation and target image.
The input images are stacked up into a 9-channel image and put through a sequence of convolutional layers to extract features. Convolving the current view with previous view will provide the knowledge of the previous action. Convolving the current view with target image indicates the signal to the goal.

\noindent\textbf{FlowNet}. We combine current view and target image into a pair and use FlowNet~\cite{dosovitskiy2015flownet} to compute the optical flow. The optical flow is represented as a two-channel flow map that is an input of our perception module.

\noindent\textbf{SiameseNet}.  The perception module
follows the architecture proposed by~\cite{agrawal2015learning}. 
It is modeled as a siamese network  for computing feature maps respectively from input images. The current observation and target image are used as inputs. The output feature maps are flattened into a vector and then concatenated into a large feature vector as the final output.

\noindent\textbf{Inverse Dynamics}. We adopt the architecture of ~\cite{pathak2018zero} to learn the network weights and train the model through supervised learning. Then we fix the weights of the trained model and use it as the perception module.

\noindent\textbf{Correlation Map}~\cite{rocco2018end}.
 The perception module computes the correlation between patches from two feature maps and outputs the correlation map as the final visual representation. The feature map extraction architecture is similar to our perception module but having two more pooling layers. We compute the correlation between two $8 \times 8 \times 64$ feature maps and obtain a $8 \times 8 \times 8 \times 8$ correlation map.

%\noindent\textbf{Overlap Area} We suspect the action decision is made through comparing the overlapping area between current observation and target image. To compute the overlapping area, we firstly compute the correspondence between the two images through known camera poses. Then for each pixel in the current view having a correspondence, we encode it with its correspondence's coordinates in the target view. The overlapping map is put through our perception module to extract features.

We separate the test cases into short-range navigation where the target location is within 10 steps starting from the initial pose and longer-range navigation where the agent needs at least 15 steps to reach the goal. It takes the robot more than 25 steps to reach the furthest goal.

\begin{table}[h]
\caption{Visual Representation}
\label{visual_representation}
\begin{center}
\begin{tabular}{|c|c|c|}
\hline
Approach & Short/Long-Range\\
\hline
\hline
LocomotionNet~\cite{savinov2018semi} & 26.5\% / 18.2\%\\

%FlowNet~\cite{dosovitskiy2015flownet} & NaN / NaN\\
\hline
SiameseNet~\cite{agrawal2015learning} & 14.3\% / 28.8\%\\
\hline
%inverseDynamics & 18.4\% / NaN\%\\
%\hline
%\hline
%CorrelationMap~\cite{rocco2018end} & NaN\% / NaN\%\\
%\hline
%CorrespondenceMap & 61.2\% / 45.4\%\\
%\hline
CorrespondenceMap & 85.7\% / 80.3\%\\
\hline
Smoothed CorrespondenceMap & \textbf{90.9\%} / \textbf{86.9}\%\\
\hline
\end{tabular}
\end{center}
\end{table}

Table~\ref{visual_representation} shows the experiment results. We didn't get satisfying results for some of the approaches so their results are missing from the table. 
% Considering that these approaches are trying to learn the visual representation while our method just take the ground-truth correspondence, it is not surprising that our method performs better. 
The optical flow detected by FlowNet is not very helpful because FlowNet is trained to learn small displacement while our displacement spans over 80 pixels in some test cases. In the following experiment, we use smoothed correspondence map as the default input when we evaluate our LVS model.  

\subsection{Reward}
Here we hold up the input visual representation to be the dense correspondence map, but vary the reward structure. 
The dense correspondence which we denote as ground-truth correspondence is computed for each pixel in the common field of view using the available depth information. 
We compare the performance of using two distance metric, $d_{polar}$ distance and $d_{pose}$ distance~\cite{lavalle2006planning}, and two reward setups, distance minimization reward $DistMinimize$ used by our model and Sadeghi's progressive reward $Progress$~\cite{sadeghi2019divis}. 

\begin{table}[h]
\caption{Reward Structure}
\label{reward}
\begin{center}
\begin{tabular}{|c|c|c|}
\hline
Reward & Distance Metric & Success Rate\\
\hline
\hline
DistMinimize & $d_{polar}$ & \textbf{83.5\%}\\
\hline
Progress & $d_{polar}$ & 73.2\%\\
\hline
Progress & $d_{pose}$ & 54.9\%\\
\hline
\end{tabular}
\end{center}
\end{table}

Table~\ref{reward} shows the results. Using both $d_{pose}$  metric and using $DistMinimize$ reward boosts the success rate for more than $10\%$. The advantage of $DistMinimize$ over $Progress$ is that in the latter, the robot will receive a positive reward even though the distance to the goal is not shortened. This makes the agent hesitant to move towards the goal. Furthermore, using $d_{pose}$ distance forces the robot to head towards the goal and avoids the risk of losing correspondence (overlap) to the target view. This is helpful for some difficult test case as shown in Figure~\ref{fig:missingCorr}.

%We hold the whole DQN training architecture but varies the distance metric when we compute the reward. Suppose the current pose is $(x_t, y_t, \theta_t)$ and the goal pose is $(x_g, y_g, \theta_g)$.
%Different distance metric largely affects the performance of the model based on our experiments. We tried three different distance metrics.Euclidean distance is the 2d distance between the two positions.
%$$\rho(x_t, y_t, x_g, y_g)=\sqrt{(x_g - x_t)^2+(y_g - y_t)^2}$$

%We designed a progressive reward to direct the agent to the target location. The reward function is $R=\max \left(0,1-\frac{\min \left(d_{t}, d_{init}\right)}{d_{init}}\right)$, where $d_init$ is the distance between start pose and target pose and $d_{t}$ is the distance between current pose and target pose. 

%Pose distance also includes in the comparison of heading angles~\cite{lavalle2006planning}.
%$$\rho(\theta_t, \theta_g)=\sqrt{(\cos\theta_g - \cos\theta_t)^2 + (\sin\theta_g - sin\theta_t)^2}$$
%$$\rho(x_t, y_t, \theta_t, x_g, y_g, \theta_g)=\rho(x_t, y_t, x_g, y_g)+\lambda \rho(\theta_t, \theta_g)$$
%Pose distance will enforce the agent's final pose has a similar heading angle as the target pose. In practice, $\lambda$ is set up to be 0.3. 

\subsection{Classical Visual Servoing Evaluation}
Here we evaluate IBVS on a non-holonomic robot from two aspects: image point features and depth data. We experiment with three point feature variations: ground-truth sparse correspondence (GtCorresp), learned sparse correspondence (LearnedCorresp) and SIFT. For ground-truth correspondence, we deliberately select $4$ correspondences with the largest offset, as these are the feature points most indicative of the motion between images. We don't evaluate dense correspondence since the interaction matrix inverse computation is time-consuming and it is more likely that the feature velocities are inconsistent when you have a large number of features. We take an off-shelf learned correspondence~\cite{detone2018superpoint} method as input. Falsely detected correspondence are removed through geometric verification, both in case of learned correspondences and SIFT features. We try four depth variations: ground-truth depth (GtDepth), constant depth (ConstantDepth), noisy depth (NoisyDepth) and no depth. Ground-truth depth image is taken directly form the simulated environment. Constant depth is setup to be $4.0$ as all the target viewpoints are within $4$ meters. Noisy depth is computed by adding a gaussian noise with $\mu=0, \sigma=0.5$ to the ground-truth depth image. For the no depth setup, we only compute the angular velocity $\omega_y$ and use constant velocity $\nu = 0.1$.

\begin{table}[h]
\caption{Different IBVS Setups}
\label{classical_VS}
\begin{center}
\begin{tabular}{|c|c|c|c|c|}
\hline
Feature+Depth & Textured & Nontextured & Corridor\\
\hline
\hline
GtCorr+GtDepth & 80.4\% & 72.2\% & 80.4\% \\
\hline
LearnedCorr + GtDepth & 40.0\% & 23.2\% & 32.9\%\\
\hline
SIFT + GtDepth & 41.3\% & 18.5\% & 24.7\%\\
\hline
GtCorr + noisyDepth & 52.0\% & 42.5\% & 46.8\%\\
\hline
GtCorr + constantDepth & 57.0\% & 56.2\% & 48.3\%\\
\hline
GtCorr + noDepth & 32.8\% & 29.9\% & 28.2\%\\
\hline
GtDenseCorr + noDepth & \textbf{87.6\%} & \textbf{88.8\%}  & \textbf{84.8}\%\\
\hline
\end{tabular}
\end{center}
\end{table}

\begin{figure}[t!]
\centering
\includegraphics[width=0.48\textwidth]{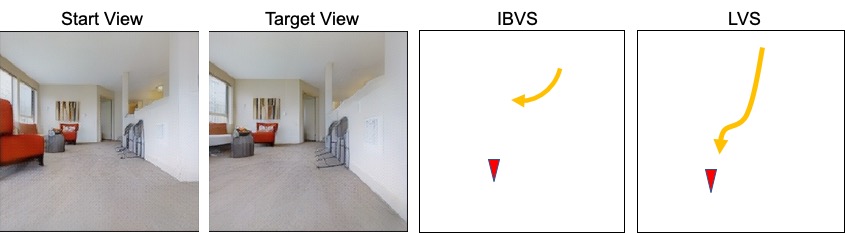}
\caption{Visualization of a test case where correspondence is missing during navigation. Given the start view and goal view, the robot should move right then turn left. Using IBVS, robot stops due to loss of FOV overlap. Using LVS, robot is able to reach the goal.}
\label{fig:missingCorr}
\end{figure}

Table~\ref{classical_VS}, compares the success rate of different IBVS methods under various environments. The results show that classical visual servoing's performance is highly correlated with the quality of feature points and depth data. Using carefully selected ground-truth correspondences boosts the model's performance by more than $40\%$ than using detected features. Using constant depth instead of ground-truth depth degrades the performance by $20\%$. Using noisy depth results in $28\%$ drop in success rate demonstrating that IBVS is sensitive to the depth changes. We also evaluate our LVS method (GtDenseCorr + noDepth) on the same test environment. It outperforms IBVS by at least $8\%$ in textured and nontextured environments especially in the following case: when the target location is on the opposite of the robot's orientation, the robot might lose the common area between current observation and the goal image. In another word, there is no correspondence between $I_t$ and $I_g$. A robot using IBVS will stop under this condition while it will continue move forward using LVS based on its learning experience. Figure~\ref{fig:missingCorr} demonstrates this hard case. %The large margin of applying LVS to non-textured environment is because there is no furniture in the scene so that fewer collisions incurred. 

\subsection{Noisy Correspondences}
% Until now we are comparing visual representations built on ground truth correspondence. It will be over optimistic when we apply our model to real environment and expect such high-quality features. 
To evaluate the robustness of our trained model we add gaussian noise to the displacement at each pixel in the correspondence map. The variance $\sigma$ of the gaussian distribution controls the amount of offset. We also vary the  density of the detected features using uniform distribution. Parameter $Coverage$ controls the probability of a pixel having a correspondence. Before we input the noisy correspondence map to the model, we convolve the map with averaging smoothing filter.  

\begin{figure}[t!]
\centering
\includegraphics[width=0.48\textwidth]{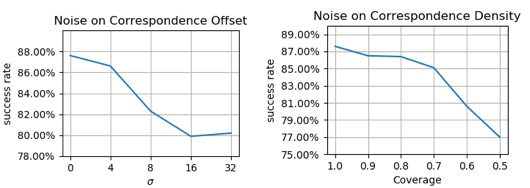}
\caption{Comparison of the performance of the model under different noise settings. Left image demonstrates the Gaussian noise on correspondence offset. Right image demonstrates the uniform noise on correspondence coverage}
\label{fig:noise}
\end{figure}

%Table~\ref{noise} 
Figure~\ref{fig:noise} shows the results. Our model is robust to large offset errors and still achieves 80\% success rate when the offset deviates from the ground-truth for up to 32 pixels. For the missed correspondence noise, it is surprising that the performance is still above $75\%$ when $50\%$ of features is missing. The smoothing reduces the sparsity and has favorable effect on the resulting policy. This demonstrates that our model is robust to several types of correspondence errors. 

\iffalse
\begin{table}[h]
\caption{Noise Affection}
\label{noise}
\begin{center}
\begin{tabular}{|c||c|}
\hline
$\sigma=0.0, s=1.0$ & 87.6\%\\
\hline
$\sigma=4.0, s=1.0$ & 86.6\%\\
\hline
$\sigma=8.0, s=1.0$ & 82.3\%\\
\hline
$\sigma=16.0, s=1.0$ & 79.9\%\\
\hline
$\sigma=32.0, s=1.0$ & 80.2\%\\
\hline
$\sigma=0.0, s=0.9$ & 86.5\%\\
\hline
$\sigma=0.0, s=0.8$ & 86.4\%\\
\hline
$\sigma=0.0, s=0.7$ & 85.1\%\\
\hline
$\sigma=0.0, s=0.6$ & 80.6\%\\
\hline
$\sigma=0.0, s=0.5$ & 77.0\%\\
\hline
\end{tabular}
\end{center}
\end{table}
\fi

\subsection{Target Object Servoing}
Even though our model is not trained on object targets, it can be easily adapted to navigate towards semantic targets. To identify the target of interest, we run an object detector~\cite{massa2018mrcnn} on the images collected by randomly driving the robot in the environment and crop the detected object of interest. The crop is then used as the target image, followed by computation of the correspondence map between current view and target view. The correspondence map is input to the model and we drive the robot by following the action predictions. Figure~\ref{fig:GoToView_and_GoToObj} gives out one example of a robot driving to an object. This suggests that the proposed method might be suitable for semantic target navigation considered previously~\cite{sadeghi2019divis}. We don't do large-scale quantitative evaluations of our model's performance on the driving-to-object task as the object locations are not labeled in the environment.

\section{CONCLUSION}
% Reaching a desired location or target is an important local navigation skill for mobile agents and can be view as an instance of visual servoing problem studied in the past. 
We demonstrated learning view and target invariant visual servoing for local navigation. 
% This entails the novel choice of the architecture, the input and reward structure for training the model. 
We examine the performance of a variety of input representations and train the model using deep Q-learning framework. Both the input of our model and the reward structure are motivated by classical visual servoing methods. The ability to train the model in an end-to-end fashion significantly improves the robustness and performance of the approach compared to classical visual servoing methods, where a small set of fixed features is selected for the task. 
We present comprehensive comparison of the model to alternative representations proposed in the literature. While the current model assumes dense correspondence, it is robust to errors in the correspondence maps. In the future work we plan to incorporate depth information explicitly, carry our more extensive experiments with target objects and presence of obstacles and validate our approach on mobile robot platform. 

\bibliographystyle{IEEEtran}
\bibliography{IEEEfull,Bibliography}

\end{document}